\documentclass[conference]{IEEEtran}
\IEEEoverridecommandlockouts
\usepackage{cite}
\usepackage{amsmath,amssymb,amsfonts}
\usepackage{algorithmic}
\usepackage{graphicx}
\usepackage{textcomp}
\usepackage{xcolor}
\def\BibTeX{{\rm B\kern-.05em{\sc i\kern-.025em b}\kern-.08em
    T\kern-.1667em\lower.7ex\hbox{E}\kern-.125emX}}

\begin{document}

\title{A Survey on Face Recognition Systems\\
}

\author{\IEEEauthorblockN{1\textsuperscript{st} Jash Dalvi}
\IEEEauthorblockA{\textit{Computer Engineering Department} \\
\textit{KJSIEIT}\\ 
Mumbai, India \\
jash.dalvi@somaiya.edu}
\and
\IEEEauthorblockN{2\textsuperscript{nd} Sanket Bafna}
\IEEEauthorblockA{\textit{Computer Engineering Department} \\
\textit{KJSIEIT}\\
Mumbai, India \\
sanket.bafna@somaiya.edu}
\and
\IEEEauthorblockN{3\textsuperscript{rd} Devansh Bagaria}
\IEEEauthorblockA{\textit{Computer Engineering Department} \\
\textit{KJSIEIT}\\
Mumbai, India \\
devansh.bagaria@somaiya.edu}
\and
\IEEEauthorblockN{4\textsuperscript{th} Mrs. Shyamal Virnodkar}
\IEEEauthorblockA{\textit{Asst. Professor, Computer Engineering Department} \\
\textit{K. J. Somaiya Institute of Engineering and Information Technology}\\
Mumbai, India \\
shyamal@somaiya.edu}
}

\maketitle

\begin{abstract}
Face Recognition has proven to be one of the most successful technology and has impacted heterogeneous domains. Deep learning has proven to be the most successful at computer vision tasks because of its convolution-based architecture. Since the advent of deep learning, face recognition technology has had a substantial increase in its accuracy. In this paper, some of the most impactful face recognition systems were surveyed. Firstly, the paper gives an overview of a general face recognition system. Secondly, the survey covers various network architectures and training losses that have had a substantial impact. Finally, the paper talks about various databases that are used to evaluate the capabilities of a face recognition system.
\end{abstract}


\section{Introduction}
Facial recognition\cite{b41} system is a technology that can efficiently and accurately detect an individual using their facial features. This system can be used to detect them in images, videos or in real-time.Traditional approaches, such as filtering responses, histograms of feature codes, and distribution of dictionary atoms, sought to distinguish human faces using one or two layer representations. Deep learning methods, like CNN, use a cascade of multiple layers of processing units for feature extraction and transformation.

DeepFace\cite{b7} achieved SOTA accuracy on the renowned LFW benchmark\cite{b44} in 2014, for the first time approaching performance level on the unconstrained situation (DeepFace: 97.35 percent vs. Human: 97.53 percent), by training a 9-layer neural network.

4 million face photos were used to create a layer model.The Gabor\cite{b43} characteristic discovered by human scientists with years of expertise is analogous to the first layer of the deep neural network. The second layer is responsible for learning more complicated texture details. The third layer's characteristics are more complicated, and certain rudimentary structures, such as a high-bridged nose and big eyes, have begun to grow. In the fourth, the network output is sufficient to describe a particular facial characteristic, which would have a discrete reaction to unambiguous abstract ideas such as smile, yell, or even blue eye.

The main aim of the introduction of face recognition is the need for security of information or physical property. In today’s advanced world, different types of frauds are taking place all over the world, using different types of security breaches. The only way of authentication was keys, ID cards, PIN numbers, etc to determine who the person is. With the introduction of an efficient facial recognition system, the system becomes extremely secure and robust, and eliminates the need to remember passwords or PIN numbers.This gives a conclusion that lower layers of CNN automatically learn the features, and higher levels are more related to a certain level of abstraction that the models learn.

\section{Overview}

Facial recognition systems\cite{b1} are the way of machines\cite{b3} getting closer to humans, making the world a step closer to automation. While facial recognition is quite easy for humans to process, it is quite the challenging part to make machines understand facial features and compute them for future recognition. There are many features that affect the facial recognition process, namely the lighting condition, face angle, how much the face is covered, different facial features like beard, mustache, spectacles, and so on. Taking these features into account, it further takes the challenge of detecting a face by machine a level ahead. 

To locate faces in photos or videos, a face detector is first utilized. The faces are aligned to normalised canonical coordinates using the facial landmark detector. These aligned face photos are used to create the Facial Recognition module. Also, before the features are fed to the facial recognition module, it is made sure that the face is live and not spoofed. For the Facial Recognition system to work, it requires three modules namely Face Processing, Deep Feature Extraction and Training Loss for Face Matching, which will be explained briefly below.

\begin{enumerate}
\item Face Pre-Processing: As mentioned earlier, facial recognition is affected by different conditions\cite{b2} like illumination, occlusions, poses, etc. To address this problem face processing is a necessary step in this system.
    
\item Deep Feature Extraction: The network architecture can be called as the backbone of the whole face recognition system\cite{b6}. Deep feature extraction is critical for multi-spectral image classification, and it's becoming a hot topic in change detection research. Because of the affects of individual variations and lighting, certain traits that are significantly connected to changes in the face are difficult to extract. As a result, characteristics that can effectively define facial changes are urgently needed.

\item Training Loss for Face Matching: Training loss has played a vital role in improving the overall accuracy of face recognition systems. The proposed paper covers two notable loss function, Euclidean loss and Cosine loss.
\end{enumerate}

\section{Face Pre-Processing}
Data augmentation, often known as "face data augmentation," is an efficient way to compensate for limited facial training data. It's a technique for increasing the quantity of training or testing data by altering real-life or simulated virtual face samples. The two methods of processing are categorised as follows: One to many augmentation, Many to one augmentation\cite{b40}.
    \begin{itemize}
	\item One to many Augmentation\cite{b4}: To allow deep networks to learn pose invariant representations, these approaches create several patches or pictures of pose variability from a single image.
	
	\item Many to one Augmentation\cite{b5}: These approaches extract the canonical view of face pictures from one or more non frontal images, allowing FR to be done under controlled settings.
    \end{itemize}
    
Types of transformation are elaborated into two categories for generating enhanced samples:
\begin{itemize}
	\item Generic Transformation: In this type of transformation, the image is altered/transformed entirely and high-level contents are ignored. It has two types Geometric and Photometric.
	
	\item Face Specific Transformation: These approach focuses mainly on facial components and attributes, allowing them to modify age, cosmetics, posture, hairstyle and facial emotions.
    \end{itemize}
    
\section{Deep Feature Extraction}
Deep Feature Extraction is done by passing input image through the architecture of the network, and a corresponding embedding feature vector is obtained. The idea of increasing the accuracy of a face recognition system relies on either improving high-quality data or improving the network architecture and training loss. Improving the data-set is a challenging task and requires a lot of resources. Some of the giants like Facebook and Google have built face recognition data-sets that include around 10\textsuperscript{6}-10\textsuperscript{7} unique faces\cite{b1}\cite{b7}. However, academics don’t have the necessary computational resources and generally rely on improving the network architecture or training loss for further improvement in the accuracy of face recognition system. For example, the accuracy of LFW Benchmark improved from 97\% to 99.7\% due to changes in architecture or tweaks in the training loss.

\begin{figure}[htp]
\centering
{\includegraphics[width=8cm]{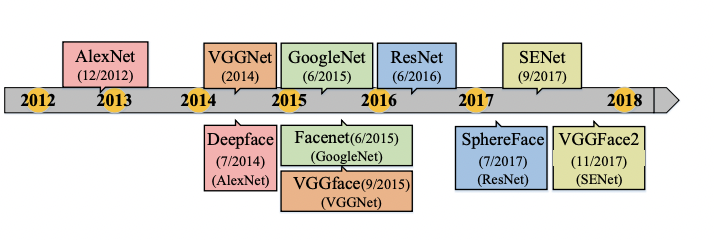}}
\caption{Evolution of Architecture}
\label{fig:database}
\end{figure}
\begin{figure}[htp]
\centering
{\includegraphics[width=8cm]{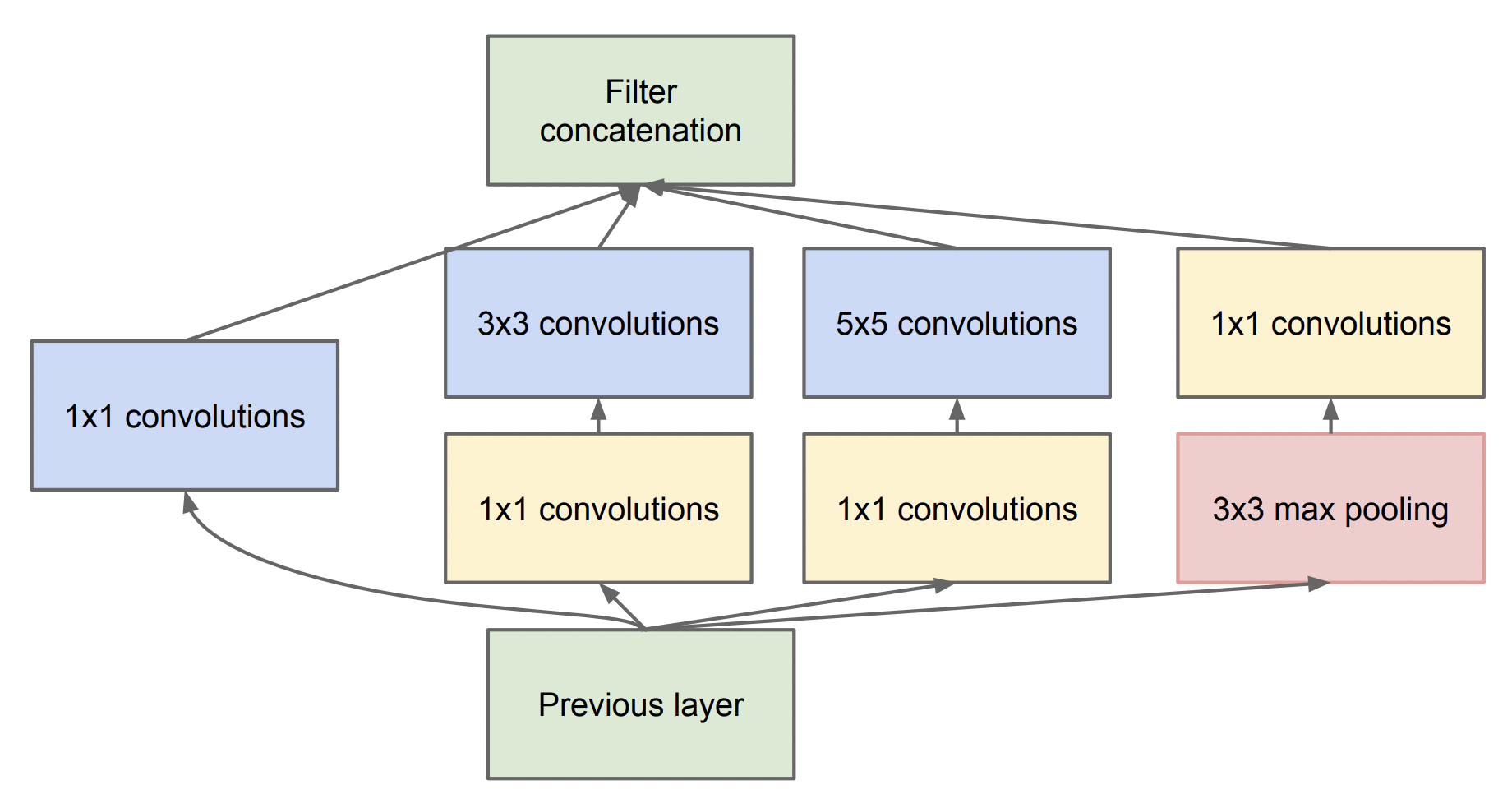}}
\caption{GoogLeNet}
\label{fig:database}
\end{figure}

The main purpose of a backbone network architecture is to extract robust features given an input facial image. The main architectures for feature extraction have always been the ones that have performed exceedingly well on general object classification problems. AlexNet\cite{b26} achieved state-of-the-art recognition accuracy in the ImageNet large-scale visual recognition competition (ILSVRC), 2012 by beating accurate models of the time by a wide margin. AlexNet is comprised of five convolution layers and three fully connected with the ReLU as the activation function, and Dropout layers and data augmentation for further increasing the performance on the test set. In 2014, VGGNet\cite{b27} proposed an architecture which comprised of 3x3 convolution kernels throughout, and additional feature maps were employed after the 2x2 max pooling layer. The depth of VGGNet waThe 22-layer GoogleNet\cite{b28} featured an "inception module" in 2015, which consisted of the concatenation of hybrid feature maps and two extra intermediate softmax activation functions. It has convolutions kernels of varying sizes(1, 1, 3, 5), and resulting feature maps are concatenated at the end of each module. Significant performance gains were also observed due to the employment of such a structure. In order to train extremely deep neural networks, perhaps of the order of around 150 layers, researchers came up with ResNet\cite{b29} model.  The key idea was the employment of a residual mapping in order to combat the earlier limitations stemming from the vanishing gradient problem. SENet \cite{b30}, the ILSVRC 2017 winner, developed a "Squeeze-and-Excitation" (SE) block that adapts to channel-wise feature responses by explicitly modeling channel inter-dependencies. These blocks may be used with current architectures like ResNet to boost their representational capability.

\section{Training loss For Face Matching}

The earliest network architectures like DeepFace\cite{b7} and DeepID\cite{b8} employed softmax loss for learning features and training the model. Researchers realized that such loss functions inherently limited the learning capability of discriminative features. Therefore, more studies began to focus on loss functions for accuracy improvements in the face recognition system. During this period, Euclidean loss and cosine loss played a vital role in the overall transformation.

\subsubsection{Euclidean Loss}
Euclidean loss\cite{b9}\cite{b10} is a learning metric that embeds a given facial image into the representative space. The basic idea behind euclidean loss is to reduce the intra-class variance between the facial images and increase the inter-class variance between facial images of different people. The most common versions of the euclidean loss are triplet loss and contrastive loss. Triplet loss \cite{b1}\cite{b11}\cite{b12}\cite{b13}\cite{b14}\cite{b15} considers the relative difference between the facial images. The equation of triplet loss is as follows:
\[ L(A,P,N) = max(|| f(A) - f(P)||^2 - || f(A) - f(N)||^2 + \alpha,0) \]
where A is known as the anchor image, P is the positive image and belongs to the same class as A, and N is the negative image which belongs to a different class than A and N. $f(.)$ is a function that converts an input image to its respective embedding vector. In constrastive loss \cite{b16}\cite{b17}\cite{b18}\cite{b19}\cite{b20}, the euclidean distance of the positive image pairs is less and the distance of negative image pairs is more. The equation of contrastive loss is as follows:
\begin{multline*}
    L = y_{ij}max(0,||f(x_{i}) - f(x_{j})||_{2} - \epsilon^+) \\
    + (1- y_{ij})max(\epsilon^- - ||f(x_{i}) - f(x_{j})||_{2},0)
\end{multline*}

where  $y_{ij}$ = 0 means $x_{i}$ and $x_{j}$ are non-matching samples and $y_{ij}$ = 1 means matching samples. f(·) is a functions that converts an image to its corresponding embedding vector, $\epsilon+$ controls the margin of matching pairs ,and $\epsilon-$ control the margin of non-matching pairs.
\subsubsection{Cosine Loss}
After the initial loss functions were developed, the researchers decided to focus on refining the loss functions in order to develop a robust face recognition system. Consequently, cosine or angular loss \cite{b21}\cite{b22}\cite{b23}\cite{b24}\cite{b25} came into the picture for further improvement of accuracy
\[L_{s} = \frac{1}{N}\sum_{i=1}^{N}-log \frac{e^{f_{y_{i}}}}{\sum_{j=1}^{C} e^{f_{j}}}\]
where N is the number of training samples and C is the number of classes. f is denoted as the activation of fully-connected layer with weight vector W.
\[f_{j} = ||W|| ||x||cos\theta_{j}\]
where $\theta$ is the angle between W and x. Both the norm and angle of vectors have an influence on the posterior probability.

\section{Face Databases}
Many face databases have been built during the last three decades, with a clear trend from small-scale to large-scale, from single-source to diverse-sources, and from lab-controlled to real-world unconstrained conditions. The face database generation procedure heavily influences the path of Face Recognition research.

\begin{figure}[htp]
\centering
{\includegraphics[width=8cm]{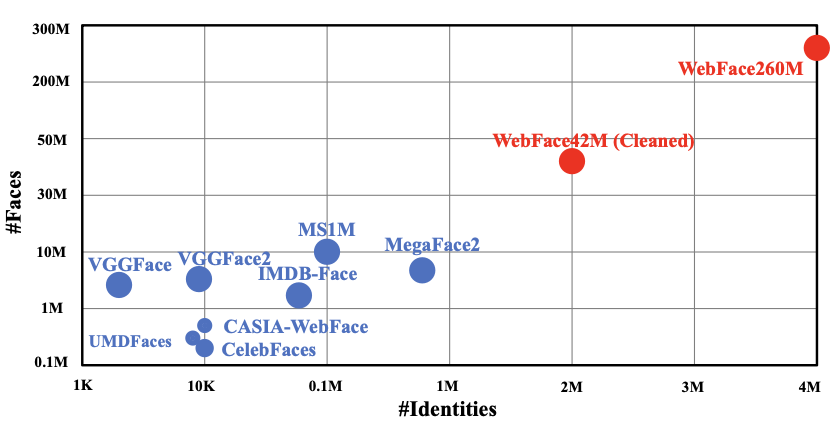}}
\caption{Number of Identities against Number of Faces.}
\label{fig:database}
\end{figure}

A sufficiently big training dataset is required for successful deep Face Recognition. A large-scale face recognition dataset is being created for the research community in order to close the data gap that exists between the industry and the research community. MS-Celeb-1M\cite{b32}\cite{b34}, VGGface2\cite{b31}, Megaface\cite{b33}\cite{b35} and WebFace260M\cite{b42} are four databases with over 1 million\cite{b38} photos. We also present an overview of some intriguing discoveries related to these training sets.

VGGFace2\cite{b31}\cite{b39},a new large-scale face dataset, contains 3.31 million photographs of 9131 persons, with an average of 362.6 shots for each subject. The photos were gathered using Google Image Search and include a variety of poses, ethnicity, ages, lighting, and jobs (e.g. actors, athletes, politicians).

The dataset was compiled with three goals in mind: (a) a large number of identities with photographs for each one; (b) a wide range of age, posture, and ethnicity; and (c) a low level of label noise. We go through how the data was collected, including the automatic and manual filtering steps that guaranteed each identity's image was accurate.

MS-Celeb-1M\cite{b34}, set a goal of recognising one million celebrities from their facial photographs and identifying them using unique entity keys in a knowledge base. The knowledge base's comprehensive information aids in disambiguation and improves recognition accuracy, as well as contributing to a variety of real-world applications like image captioning and news video analysis. Our face recognition work is practical and advantageous to many real-world applications, such as image search, ranking, caption creation, image deep understanding, and so on, thanks to the rich information.

WebFace260M\cite{b42} is a new million-scale face benchmark that comprises training data such as noisy\cite{b37} 4M identities/260M faces (WebFace260M) and cleaned 2M identities/42M faces (WebFace42M), as well as a time-constrained evaluation method.
Pre-processing of the face: Five landmarks that are utilised to recognise and align faces are predicted by Retina Face\cite{b36}. WebFace260M's statistics are displayed as Date of Birth, Nationality, and Profession. The WebFace260M training set is automatically cleaned using the CAST method, and the WebFace42M training set is created. The face attribute annotations in WebFace42M include posture, race, age, hat, glass, mask, and gender.

\section{Conclusion}
Face recognition technology has proven to be one of the most impactful technology in recent decades. The proposed paper covers various face recognition architectures and proceeds by covering various parameters needed for training such models. First of all, the paper starts by giving a general introduction of face recognition and its ramifications. Secondly, it gives a general overview of the face recognition pipeline and starts with elaboration of first aspect, face pre-processing. Further, the paper talks about various network architectures employed for deep feature extraction and different loss functions used to train such models and also for face matching. Lastly, it gives information about various face identification databases that are used by the academic community to compare the accuracy metric with respect to other results proposed by other contemporary studies. The paper is an effort to summarize notable advances in face recognition technology in the last decade.

\end{document}